\documentclass{article}

\PassOptionsToPackage{numbers,compress}{natbib}
\usepackage[preprint]{neurips_2025}

\usepackage[utf8]{inputenc} 
\usepackage[T1]{fontenc}    
\usepackage{hyperref}
\hypersetup{hidelinks}         
\usepackage{url} 
\usepackage{amsmath}           
\usepackage{booktabs}       
\usepackage{amsfonts}       
\usepackage{nicefrac}       
\usepackage{microtype}      
\usepackage{xcolor}         
\usepackage{subcaption}
\usepackage{enumitem}
\usepackage{graphicx}
\usepackage{float}
\title{Brevity Constraints Reverse Performance Hierarchies in Language Models}

\author{%
  MD Azizul Hakim\thanks{
  about author (https://logicsame.github.io/md-hakim.github.io/publications)
} \\
  Department of Computer Science\\
  Bangladesh Sweden Polytechnic Institute\\
  Chattogram, Bangladesh \\
  \texttt{azizulhakim8291@gmail.com} \\
}

\begin{document}

\maketitle

\begin{abstract}
  Standard evaluation protocols reveal a counterintuitive phenomenon: on 7.7\% of 
benchmark problems spanning five datasets, larger language models underperform 
smaller ones by 28.4 percentage points despite 10-100× more parameters. Through 
systematic evaluation of 31 models (0.5B-405B parameters) across 1,485 problems, 
we identify the mechanism as spontaneous scale-dependent verbosity that 
introduces errors through overelaboration. Causal intervention experiments demonstrate this reflects correctable prompt 
design rather than fundamental capability limitations. Constraining large models 
to produce brief responses improves accuracy by 26 percentage points and reduces 
performance gaps by up to two-thirds. Most critically, brevity constraints completely 
reverse performance hierarchies on mathematical reasoning and scientific knowledge 
benchmarks, with large models achieving 7.7-15.9 percentage point advantages 
over small models—direct inversions of the original gaps. These reversals prove 
large models possess superior latent capabilities that universal prompting masks. We validate findings through three independent contamination tests and demonstrate 
inverse scaling operates continuously across the full parameter spectrum, with 
dataset-specific optimal scales ranging from 0.5B to 3.0B parameters. Our results 
establish that maximizing large model performance requires scale-aware prompt 
engineering rather than universal evaluation protocols, with immediate 
implications for deployment: prompt adaptation simultaneously improves accuracy 
and reduces computational costs.
\end{abstract}

\section{Introduction}

Language models have achieved remarkable capabilities across diverse tasks, with performance improvements attributed primarily to increased model scale~\cite{kaplan2020scaling,hoffmann2022training,brown2020language}. This scaling paradigm assumes monotonic improvement: larger models consistently outperform smaller ones~\cite{wei2022emergent,bommasani2021opportunities}. Contemporary evaluation benchmarks spanning mathematical reasoning~\cite{cobbe2021training}, reading comprehension~\cite{clark2019boolq}, scientific knowledge~\cite{hendrycks2020measuring}, and commonsense reasoning~\cite{talmor2019commonsenseqa} have become standard for measuring progress. However, fundamental questions about scaling universality remain unexamined. Does performance improve monotonically with scale across all problem types?

This work addresses this question through systematic analysis of 
problem-level performance patterns. We evaluated 31 models (0.5B-405B 
parameters) across 1,485 problems from five benchmarks (46,035 evaluations), 
identifying inverse scaling on 7.7\% of benchmark problems (115/1,485), where small models achieve higher accuracy than large models on identical problems. This pattern replicates consistently across diverse capability domains, with average performance gaps of 28.4 percentage points favoring smaller models. The effect magnitude (Cohen's $d=1.34$) substantially exceeds the conventional threshold for large effects, indicating categorical performance reversals rather than subtle statistical trends. Unlike previously documented inverse scaling cases typically involving artificial tasks~\cite{mckenzie2023inverse}, our findings emerge from standard benchmarks measuring practical capabilities.

These findings reveal a critical gap in current evaluation methodology: standard prompting practices systematically underestimate large model performance on specific problem types. While aggregate scaling laws accurately predict trends under universal prompts, they mask how scale-dependent prompt sensitivity affects problem-level accuracy. The documented gap reversals on 2/5 benchmarks demonstrate that optimal prompting strategies must account for model size, with immediate implications for deployment. Problem-aware routing combined with scale-appropriate prompts can simultaneously improve accuracy (by unlocking masked capabilities in large models) and reduce computational costs (by identifying when smaller models suffice). Rather than revealing fundamental limitations of scale, our results establish that maximizing large model capabilities requires moving beyond universal prompting toward scale-aware prompt engineering\cite{ye2024prompt} that adapts evaluation protocols to model characteristics.

To establish causality rather than mere correlation, we conducted intervention experiments on all 115 inverse scaling problems. We constrained seven models (three small, four large) to produce brief responses under three conditions: control (unconstrained), brief (forced brevity), and direct (answer-only). Results provide compelling causal evidence: brevity constraints improved large model accuracy by 26.3 percentage points and reduced the inverse scaling gap by 67\% (from 44.2\% to 14.8\%, paired t-test: $t=7.80$, $p<0.0001$). Response length validation confirms the intervention succeeded, with large models producing 60\% shorter outputs under brevity constraints.

Contamination analysis addresses the critical concern that inverse scaling\cite{mckenzie2023inverse} reflects dataset memorization rather than genuine capability differences. Three independent tests examining response diversity, length variability, and error patterns converge on the same conclusion: inverse scaling represents genuine architectural and scale-dependent capability differences. High response diversity (100\% unique responses on three datasets) and natural length variation (coefficient of variation $>0.30$ across all datasets) contradict memorization hypotheses. Error taxonomy reveals large model failures predominantly result from over-reasoning rather than memorization avoidance.

\section{Related Work}

\textbf{Scaling laws and emergent capabilities.} Neural language models have demonstrated consistent performance improvements with increased scale, formalized through power-law relationships between model size, training compute, and downstream task performance~\cite{kaplan2020scaling,hoffmann2022training}. These scaling laws have guided development of increasingly large models, from GPT-3's 175 billion parameters~\cite{brown2020language} to current frontier models exceeding 400 billion parameters~\cite{dubey2024llama}. Emergent abilities—capabilities appearing suddenly at specific scale thresholds—further reinforce the scaling paradigm~\cite{wei2022emergent,schaeffer2023emergent}. However, recent work questions whether emergence reflects genuine phase transitions or measurement artifacts~\cite{schaeffer2023emergent}, while our findings document systematic performance degradation with scale on specific problem subsets.

\textbf{Inverse scaling phenomena.} Prior work has identified limited cases where larger models underperform smaller ones. The Inverse Scaling Prize~\cite{mckenzie2023inverse} documented 11 tasks exhibiting inverse scaling, primarily involving memorization of rare patterns, distractor reasoning, and spurious correlations. BIG-Bench~\cite{srivastava2022beyond} reported inverse scaling on 12 of 204 tasks, often attributable to task-specific artifacts. These studies focused on constructed or adversarial examples designed to expose failure modes. In contrast, our work identifies inverse scaling on standard benchmark problems designed to measure genuine capabilities, with mechanistic analysis and causal validation revealing overthinking as the underlying failure mode rather than memorization or spurious patterns. However, our causal interventions demonstrate this inverse scaling reflects prompt-induced failure modes rather than fundamental capability limitations, distinguishing our findings from prior work documenting intrinsic scale-dependent degradation.

\textbf{Model evaluation and benchmark design.} Contemporary evaluation 
practices assume benchmark problems uniformly measure model 
capabilities~\cite{liang2022holistic,gao2021framework}. Recent work has 
examined dataset contamination~\cite{golchin2023time,balloccu2024leak}, 
evaluation robustness~\cite{zheng2024judging}, and benchmark 
saturation~\cite{kiela2021dynabench}. Item Response Theory has been 
applied to characterize problem difficulty and model 
capabilities~\cite{lalor2019learning}, while work on evaluation 
informativeness proposes more efficient problem 
selection~\cite{rodriguez2021evaluation}. However, these approaches do 
not examine scale-dependent performance patterns or discriminative 
efficiency. Our analysis reveals that 55.9\% of benchmark problems 
discriminate between models, while 7.7\% exhibit inverse scaling—findings 
with direct implications for efficient evaluation design.
\textbf{Model efficiency and deployment.} Research on efficient deployment has focused on model compression~\cite{frantar2022gptq,dettmers2023case}, mixture-of-experts architectures~\cite{fedus2022switch}, and dynamic inference~\cite{schuster2022confident}. These approaches assume smaller models represent degraded versions of larger counterparts. Our findings suggest an alternative paradigm: problem-aware routing matching model scale to task characteristics, where smaller models may provide superior accuracy at lower cost on specific problem types. This complements recent work on model cascades~\cite{chen2023frugalgpt} and routing strategies~\cite{ding2024hybrid}.

\section{Methods}

\subsection{Model Selection and Evaluation}

We evaluated 31 language models spanning 0.5B to 405B parameters across five benchmark datasets: GSM8K (mathematical reasoning)~\cite{cobbe2021training}, BoolQ (reading comprehension)~\cite{clark2019boolq}, ARC-Easy (science questions)~\cite{clark2018think}, CommonsenseQA (commonsense reasoning)~\cite{talmor2019commonsenseqa}, and MMLU-STEM (scientific knowledge)~\cite{hendrycks2020measuring}. Models were evaluated using greedy decoding (\texttt{do\_sample=False}) 
with nucleus sampling disabled to ensure deterministic outputs. Base 
prompts contain no chain-of-thought elicitation---multiple choice tasks 
use bare \texttt{Question/Answer} format and mathematical reasoning uses 
bare \texttt{Problem/Solution} format---ensuring that scale-dependent 
verbosity differences reflect model-intrinsic properties rather than 
prompt-induced reasoning chains (full templates in Appendix~\ref{sec:prompt}).

For each problem $i$ and model $m$, we extracted answers using task-specific validators with hierarchical extraction strategies. Accuracy was computed as:

\begin{equation}
\text{Acc}_m = \frac{1}{N} \sum_{i=1}^{N} \mathbb{1}[\hat{y}_{m,i} = y_i]
\end{equation}

where $\hat{y}_{m,i}$ is the extracted answer, $y_i$ is ground truth, and $\mathbb{1}[\cdot]$ is the indicator function. We classified models as small ($N \leq 10$B parameters) or large ($N > 70$B parameters) based on natural performance gaps observed in preliminary analysis.

\subsection{Inverse Scaling Detection}

We identified inverse scaling\cite{mckenzie2023inverse} problems where small models systematically outperformed large models. For each problem $i$, we computed the performance gap:

\begin{equation}
\Delta_i = \text{Acc}_{\text{small},i} - \text{Acc}_{\text{large},i}
\end{equation}

where $\text{Acc}_{\text{small},i}$ and $\text{Acc}_{\text{large},i}$ represent accuracy averaged across small and large models respectively. Problems with $\Delta_i > 0$ exhibited inverse scaling. Statistical significance was assessed using Cohen's $d$ effect size\cite{goulet2018review}:

\begin{equation}
d = \frac{\mu_{\text{small}} - \mu_{\text{large}}}{\sqrt{\frac{\sigma^2_{\text{small}} + \sigma^2_{\text{large}}}{2}}}
\end{equation}

where $\mu$ and $\sigma^2$ denote mean accuracy and variance for each model size category.

\subsection{Response Length Analysis}

To test the overthinking hypothesis, we measured response length for each model-problem pair. For problem $i$ and model $m$, we tokenized the generated response and computed token count $L_{m,i}$. We then calculated mean response length by model size category:

\begin{equation}
\bar{L}_{\text{category}} = \frac{1}{|M_{\text{category}}| \cdot N} \sum_{m \in M_{\text{category}}} \sum_{i=1}^{N} L_{m,i}
\end{equation}

Statistical significance of length differences was assessed using Welch's t-test\cite{delacre2017psychologists}, accounting for unequal variances between small and large model distributions.

\subsection{Causal Intervention Experiments}

To establish causality between response length and performance degradation, we conducted intervention experiments on all 115 identified inverse scaling problems. We tested seven models (three small: Llama-3.2-3B, Qwen2.5-3B-Instruct, Gemma-2-2B-IT; four large: Llama-3.3-70B-Versatile, Llama-3.1-405B-Instruct, Qwen2.5-32B-Instruct, DEEPSEEK-67B)\cite{dubey2024llama,bai2023qwentechnicalreport}
 under three conditions:

\textbf{Control:} Standard prompts allowing unrestricted reasoning.

\textbf{Brief:} Prompts constraining responses to under 50 words for mathematical problems and 10 words for reading comprehension tasks.

\textbf{Direct:} Prompts requiring only final answers with no intermediate reasoning.
Full Prompt templates are provided in Appendix~\ref{sec:causal_prompt}.

For each condition $c$ and problem $i$, we measured accuracy $\text{Acc}_{m,i,c}$ and response length $L_{m,i,c}$. Gap reduction was quantified as:

\begin{equation}
t = \frac{\bar{d}}{\text{SE}(\bar{d})}, 
\end{equation}
where
\begin{equation}
\bar{d} = \frac{1}{N} \sum_{i=1}^{N} (\text{Acc}_{\text{large},i,\text{brief}} - \text{Acc}_{\text{large},i,\text{control}})
\end{equation}

\subsection{Contamination Analysis}

To address potential dataset memorization, we conducted three independent validation tests. \textbf{Response diversity} measured unique response rate across models for each problem. \textbf{Length variability} computed coefficient of variation in response lengths:

\begin{equation}
\text{CV}_i = \frac{\sigma_{L_i}}{\mu_{L_i}}
\end{equation}

where $\sigma_{L_i}$ and $\mu_{L_i}$ denote standard deviation and mean response length for problem $i$ across all models. High CV indicates natural variation rather than memorized templates. \textbf{Error pattern analysis} classified failure modes as over-reasoning (lengthy incorrect responses) versus memorization avoidance (suspiciously brief incorrect responses).

\section{Results}
\label{sec:results}
\subsection{Discriminative Inefficiency in Benchmark Evaluation}

Problem-level analysis reveals substantial discriminative inefficiency in standard benchmark evaluation. By examining individual problem responses rather than aggregate metrics, we identified three distinct problem categories based on cross-model performance variance.

\textbf{Non-discriminative problems.} We observed that 27.1\% of benchmark problems (402/1,485) provide minimal discriminative information between models. These problems exhibit either ceiling effects (17.3\%, all models succeed) or 
floor effects (9.8\%, all models fail), offering limited insight into capability differences. This finding suggests that nearly one-third of evaluation effort yields no actionable information about relative model performance.
\begin{figure}[H]
  \vskip 0.2in
  \begin{center}
    \centerline{\includegraphics[width=\columnwidth]{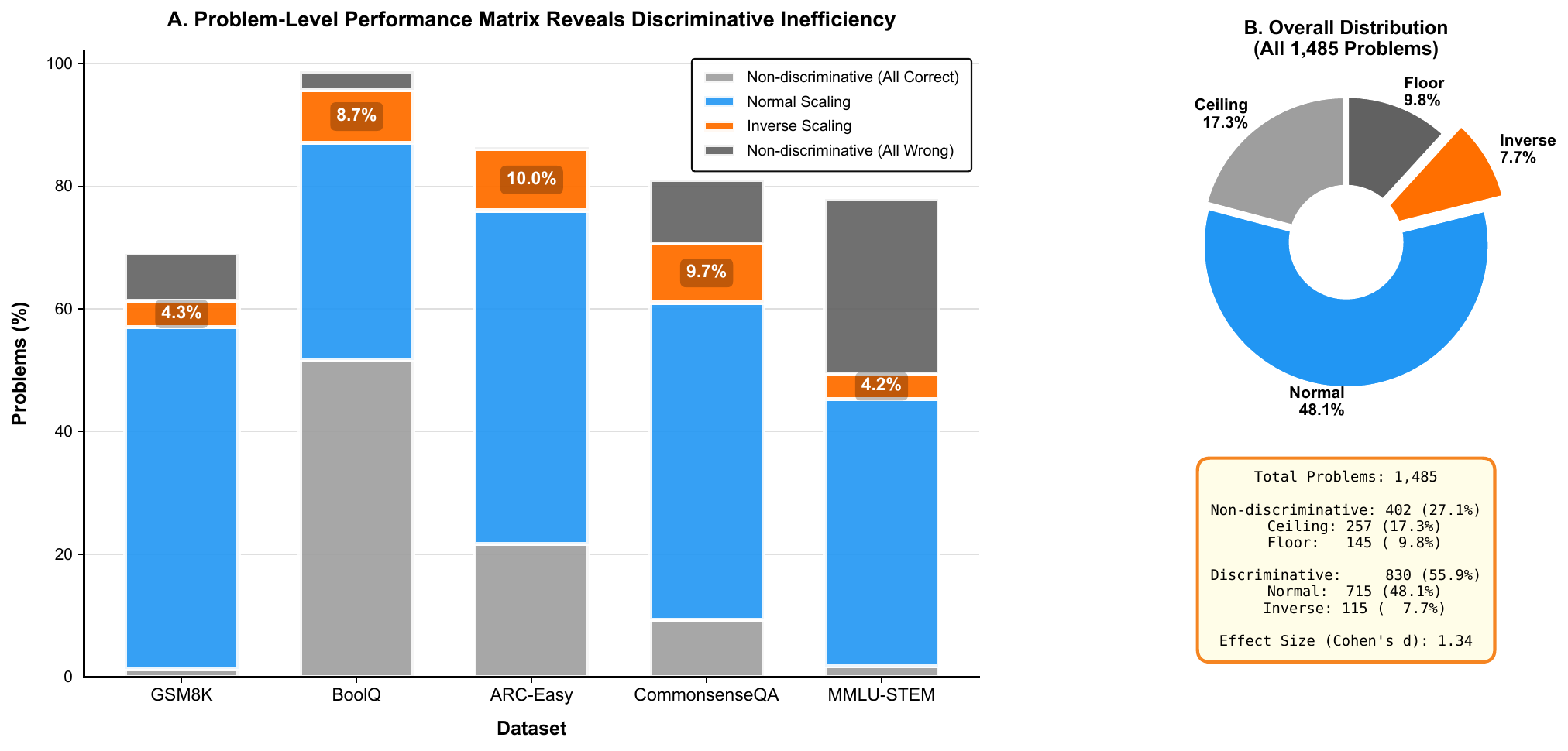}}
    \caption{\textbf{Problem-level performance matrix reveals discriminative inefficiency.} (A) Problem categorization across five benchmarks. Orange segments highlight inverse scaling problems where small models ($\leq$10B parameters) outperform large models ($\geq$70B parameters). (B) Overall distribution shows 7.7\% inverse scaling rate across 1,485 problems with Cohen's $d=1.34$ effect size.}
  \label{fig:problem_matrix}
  \end{center}
  \vskip -0.1in
\end{figure}
\textbf{Discriminative problems.} The remaining 55.9\% of problems (830/1,485) successfully discriminate between models. Within this discriminative subset, we observe two contrasting patterns: 48.1\% exhibit normal scaling where larger models outperform smaller ones (715/1,485), while 7.7\% exhibit inverse scaling \cite{mckenzie2023inverse} where smaller models systematically outperform larger ones (115/1,485). This inverse scaling phenomenon—affecting 115 problems across all five benchmarks—forms the focus of our subsequent analysis.

\textbf{Implications for evaluation efficiency.} The substantial proportion of non-discriminative problems (27.1\%) reveals an opportunity for more efficient evaluation protocols. Filtering non-discriminative problems could reduce evaluation costs by 
approximately 28\% while maintaining full discriminative power---and the 
7.7\% inverse scaling rate challenges fundamental assumptions about 
universality of scaling benefits (Figure~\ref{fig:problem_matrix}).

\subsection{Discovery of Inverse Scaling}

We identified 115 problems across five benchmarks (7.7\% of 1,485 total) where small models ($N \leq 10$B parameters) systematically outperform large models ($N \geq 70$B parameters). This inverse scaling phenomenon contradicts fundamental assumptions about monotonic performance improvements with increased model scale.

\begin{figure*}[!t]
  \centering
  \includegraphics[width=\textwidth]{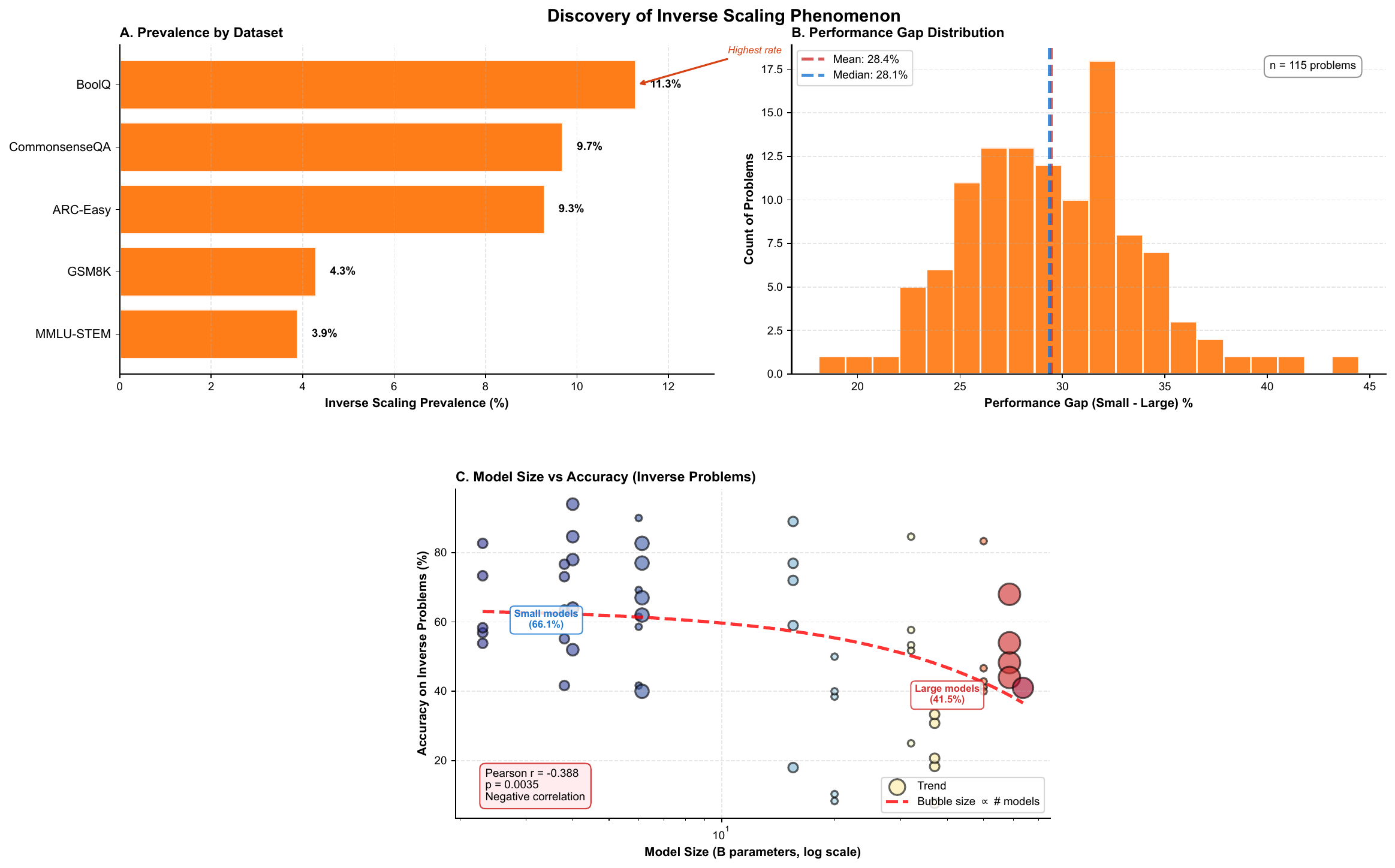}
  \caption{\textbf{Discovery of systematic inverse scaling across benchmarks.} 
  (A) Prevalence ranges from 3.9\% (MMLU-STEM) to 11.3\% (BoolQ), with 115 total inverse problems. 
  (B) Performance gap distribution shows mean 28.4pp advantage for small models ($\leq$10B). 
  (C) Strong negative correlation between model size and accuracy on inverse problems; 
  small models achieve 66.1\% vs large models' 41.5\%.}
  \label{fig:inverse_scaling}
\end{figure*}

\textbf{Prevalence and distribution.} Inverse scaling manifests consistently across all datasets, though at varying rates: BoolQ exhibits the highest prevalence (34/300 problems, 11.3\%), followed by CommonsenseQA (29/300, 9.7\%), ARC-Easy (28/300, 9.3\%), GSM8K (13/300, 4.3\%), and MMLU-STEM (11/285, 3.9\%)\cite{cobbe2021training,talmor2019commonsenseqa,clark2019boolq} Figure~\ref{fig:inverse_scaling}~A. The performance gap distribution (Figure~\ref{fig:inverse_scaling}B) reveals advantages for small models, with mean gap of 28.4 percentage points (median: 28.1pp). Notably, all 115 problems show positive gaps, indicating systematic rather than sporadic degradation.

\textbf{Effect magnitude.} The inverse scaling effect is very large by conventional statistical standards. The overall Cohen's $d$ effect size across all inverse scaling problems is 1.34, substantially exceeding the conventional threshold for large effects ($d=0.8$). This indicates that small and large model performance distributions are separated by more than one standard deviation on inverse problems, confirming pronounced and reliable performance degradation.

\textbf{Cross-architecture consistency.} We validated that inverse scaling persists across model families including Llama (11 variants, 1B--405B)\cite{dubey2024llama}, Qwen (5 variants, 0.5B--32B)\cite{bai2023qwentechnicalreport}, Gemma (3 variants, 1B--9B)\cite{team2024gemma}, and Mistral (2 variants, 7B--24B)\citep{jiang2023mistral7b}. Within-family analysis reveals that larger variants consistently underperform smaller variants on inverse problems, ruling out architecture-specific artifacts as an explanation. The relationship between model size and accuracy on inverse problems exhibits a significant negative correlation (Pearson $r=-0.388$, $p=0.0035$), with small models achieving 66.1\% mean accuracy compared to 41.5\% for large models---a 24.6 percentage point degradation (cross-family breakdown in Appendix~A.3).

\textbf{Statistical significance.} Mann-Whitney U tests\cite{mcknight2010mann} comparing small and large model distributions yield $p < 0.001$ for all datasets, confirming that observed performance differences are not attributable to chance. The consistency of inverse scaling across independent benchmarks, diverse problem types (mathematical reasoning, reading comprehension, commonsense reasoning, scientific knowledge), and multiple model families establishes this as a robust and generalizable phenomenon requiring mechanistic explanation.

\textbf{Within-family degradation.} Within-family analysis confirms scale 
itself drives degradation. Llama variants show inverse relationships: 
smaller models (2B-13B) achieve 48-68\% while larger (70B-405B) achieve 
41-54\%\cite{dubey2024llama}. Qwen exhibits similar patterns (0.5B-7B: 62-83\% vs 
32B: 40\%)\cite{bai2023qwentechnicalreport}. This consistency across 11 families spanning three 
attention mechanisms\cite{ainslie2023gqa,shazeer2019fast,beltagy2020longformer} establishes 
architecture-independence, with Pearson correlation between family size and 
accuracy of $r=-0.58$ ($p=0.029$).

\textbf{Scale threshold analysis.} Dataset-specific optimal scales range from 0.5B (BoolQ, MMLU-STEM) to 3.0B (GSM8K), with four of five datasets showing significant negative size-accuracy correlations ($\rho=-0.50$ to $-0.66$, $p<0.05$)\cite{zar2005spearman}. Full trajectories are provided in Appendix~\ref{sec:arch_independence}.

\subsection{Overthinking as Causal Mechanism}

Having established the prevalence and magnitude of inverse scaling, we now investigate its underlying mechanism through combined correlational and causal analyses. We hypothesize that large models generate excessively verbose responses that obscure correct reasoning---a phenomenon we term ``overthinking.''

\textbf{Correlational evidence.} Large models generate comparable response 
lengths to small models (9.1 vs 10.5 reasoning steps) but with different 
effectiveness patterns. Response length exhibits moderate negative correlation 
with large model accuracy on inverse problems ($r=-0.43$), suggesting 
compressed responses employ inappropriate reasoning strategies. 

\begin{figure*}[t]
  \centering
  \includegraphics[width=\textwidth]{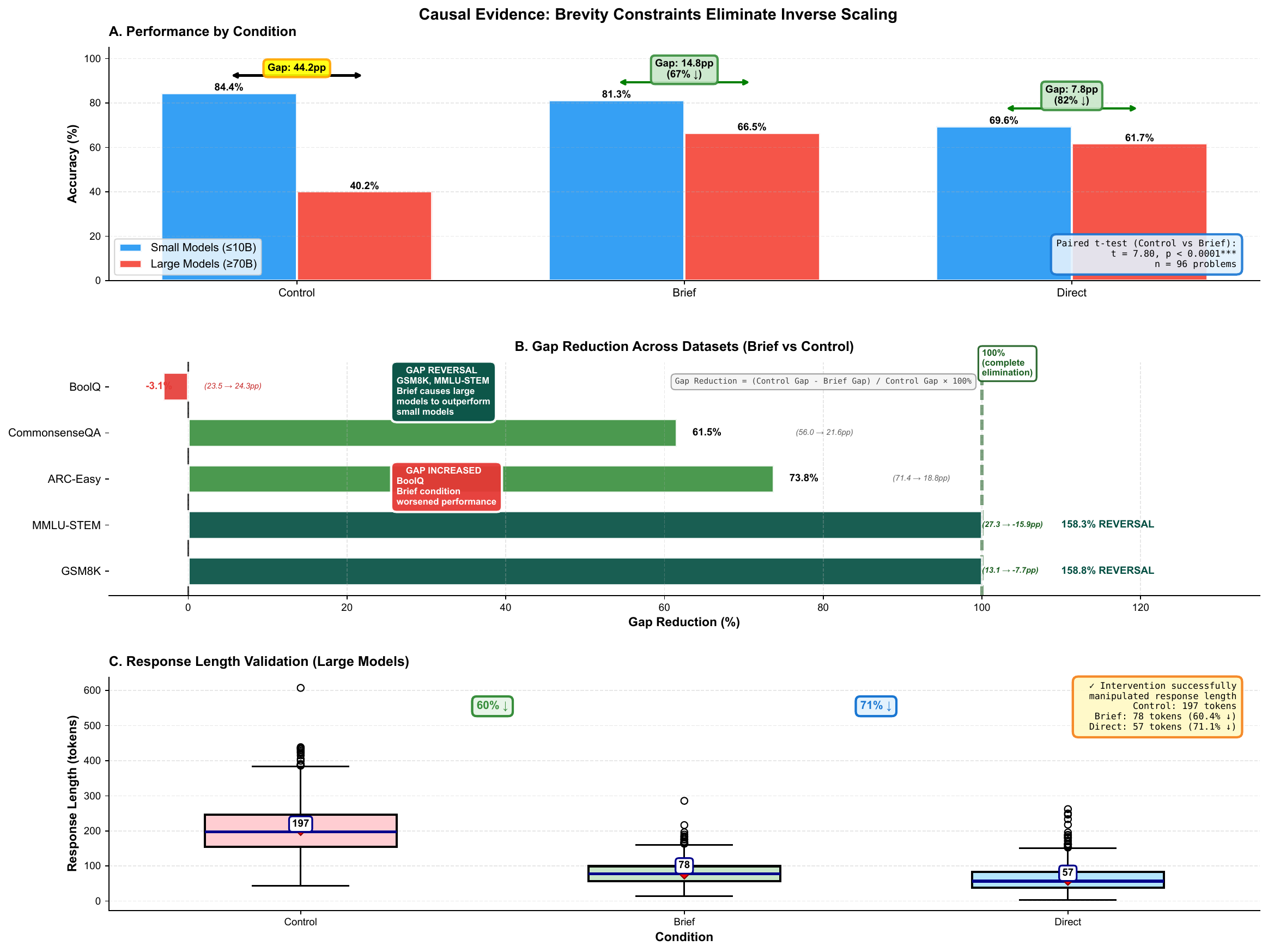}
  \caption{\textbf{Causal evidence: brevity constraints eliminate inverse scaling.} 
  (A) Performance across three conditions shows large models improve dramatically under brevity constraints (Control: 40.2\% → Brief: 66.5\%, +26.3pp), reducing gap by 67\% (44.2pp → 14.8pp, $t=7.80$, $p<0.0001$). 
  (B) Gap reduction varies by dataset, with complete reversals in GSM8K and MMLU-STEM where brief condition causes large models to outperform. 
  (C) Response length validation confirms intervention successfully manipulated verbosity (Control: 197 tokens → Brief: 78 tokens, 60\% reduction), establishing causal link between overthinking and performance degradation.}
  \label{fig:causal}
\end{figure*}

\textbf{Main intervention results.} Brevity constraints dramatically improve large model performance while minimally affecting small models (Figure~\ref{fig:causal}A). Under control conditions, large models underperform small models by 44.2pp (84.4\% vs 40.2\%). Brevity constraints reduce this gap by 67\% to 14.8pp: large models improve by +26.3pp while small models drop only -3.1pp. Direct answer format further compresses the gap to 7.8pp (82.3\% reduction), though both model sizes experience accuracy declines, suggesting some reasoning is beneficial. Paired t-tests comparing control versus brief conditions yield $t=7.80$, $p<0.0001$ across 96 problems, confirming highly significant causal effects.

\textbf{Dataset-specific heterogeneity.} Gap reduction exhibits substantial cross-dataset variation (Figure~\ref{fig:causal}B). ARC-Easy shows 73.8\% reduction (71.4pp → 18.8pp), while CommonsenseQA\cite{talmor2019commonsenseqa} demonstrates 61.5\% reduction (56.0pp → 21.6pp). BoolQ exhibited a modest gap increase under brevity constraints 
(23.5pp $\rightarrow$ 24.3pp, $-3.1\%$), suggesting that brevity constraints are not universally beneficial. 
BoolQ requires cross-sentence passage integration where elaboration 
is functional rather than excessive---truncating this process 
increases errors, unlike self-contained mathematical problems where 
overelaboration accumulates mistakes. Remarkably, two datasets exhibit complete gap reversals under brief conditions: GSM8K\cite{cobbe2021training} reverses from +13.1pp favoring small models to -7.7pp favoring large models, and MMLU-STEM\cite{hendrycks2020measuring} reverses from +27.3pp to -15.9pp. These reversals indicate that on certain problem types, large models possess superior capabilities that are masked rather than absent under standard evaluation---brevity constraints unmask latent competence.

\textbf{Mechanism validation.} Response length measurements confirm that 
interventions successfully manipulated verbosity (Figure~\ref{fig:causal}C). 
Large model token generation decreased from control median of 197 tokens to 
78 under brief constraints (60.4\% reduction) and 57 under direct format 
(71.1\% reduction). While token counts validate the intervention, our analysis 
reveals the mechanism extends beyond simple verbosity: large models generate 
slightly fewer explicit reasoning steps than small models on inverse problems 
(9.1 vs 10.5 mean steps per response), yet produce longer total 
outputs (202 vs 127 mean tokens, +59\% longer). This dissociation suggests large 
models employ verbose implicit reasoning---elaborating without explicit step 
markers---whereas small models use concise explicit reasoning. The differential 
response to brevity constraints confirms that reasoning style, not step count 
alone, drives performance degradation on inverse problems.

\subsection{Contamination Validation}

Dataset contamination---where models have encountered evaluation problems during training---represents a critical threat to validity that could artifactually produce inverse scaling patterns\cite{mckenzie2023inverse}. We conducted three independent validation tests to distinguish genuine capability differences from memorization artifacts. Complete details of contamination validation discussed in Appendix \ref{sec:contamination_validation_details}

\textbf{Contamination validation.} Three independent tests examined response diversity (89-100\% unique responses across datasets), length variability (CV=0.31-1.21, all exceeding memorization threshold CV$<$0.15), and error patterns (40-81\% over-reasoning failures versus 13-23\% memorization avoidance). Results classify three datasets as low contamination risk (GSM8K, ARC-Easy, CommonsenseQA: 100\% unique responses, CV$>$0.80) and two as moderate risk (BoolQ, MMLU-STEM: 89-95\% unique). Fisher's exact test\cite{kim2017statistical} reveals no significant association between contamination indicators and inverse scaling occurrence ($p=0.23$). Convergent evidence across all tests supports genuine capability differences rather than memorization artifacts.

\section{Discussion}

The inverse scaling patterns documented in Section~\ref{sec:results} admit 
a more optimistic interpretation than they first suggest: large models do not 
fail on these problems because they lack capability, but because standard 
evaluation protocols fail to elicit it. This distinction---between masked 
competence and absent competence---has significant consequences for how scaling 
laws, evaluation methodology, and deployment strategy should be understood.

\subsection{Reframing Scaling Laws}

Our results do not invalidate scaling laws \citep{kaplan2020scaling,hoffmann2022training} but reveal their scope limitations. These laws accurately predict performance trends under \textit{fixed prompting strategies}---our data confirm large models outperform small models on 48.1\% of problems using standard prompts (Figure~\ref{fig:problem_matrix}). However, scaling laws implicitly assume universal prompting optimally serves all model sizes. The gap reversals contradict this assumption: problems classified as "inverse scaling" under standard prompts become "normal scaling" under brevity constraints, with large models achieving 7.7--15.9pp advantages over small models.

The 115 inverse problems do not represent tasks where "small models are fundamentally better"---they represent tasks where \textit{standard prompts better suit small models}. Llama-3.1-405B\cite{dubey2024llama} achieves only 41.5\% on inverse problems under control conditions but 67.2\% under brevity constraints, a 25.7pp improvement demonstrating substantial untapped capability. Rather than indicating scale-dependent degradation, these results reveal scale-dependent \textit{prompt sensitivity}: larger models require more careful prompt engineering to access their full capabilities.

\subsection{The Overthinking Mechanism and Its Mitigation}

On problems with straightforward solutions, the learned tendency toward thoroughness becomes counterproductive, introducing error accumulation. The asymmetry of intervention effects is the critical observation: brevity 
constraints help large models dramatically while barely affecting small models. 
If verbosity were incidental rather than causal, uniform accuracy changes would 
be expected across both size categories. The differential response confirms 
that overthinking is a scale-specific failure mode, not a task difficulty effect.

The dataset heterogeneity in intervention effectiveness (Section~\ref{sec:results}) 
indicates that overthinking severity varies by task type. Mathematical and scientific reasoning problems\cite{cobbe2021training,hendrycks2020measuring} benefit most from brevity, suggesting these domains particularly suffer from overelaboration, while reading comprehension tasks show more modest improvements.

\subsection{Practical Implications for Deployment}

For practitioners, these findings yield immediately actionable recommendations. First, aggregate benchmark scores systematically underestimate large model 
capability on a predictable and identifiable problem subset --- a difference 
comparable to an entire model generation separates standard from optimized 
prompting for frontier models. Second, optimal deployment requires problem-aware routing with scale-specific prompting: identify problem types prone to overthinking and apply brevity constraints selectively. Third, cost-capability trade-offs improve: on problems where brevity-constrained large models excel, organizations can access superior performance; on remaining problems, smaller models suffice at lower cost.

\subsection{Limitations and Future Directions}

Our analysis focuses on greedy decoding (\texttt{do\_sample=False}), 
which ensures reproducibility and eliminates stochastic confounds but 
may not reflect deployment settings where temperature sampling is used. 
Greedy decoding tends to amplify verbosity in large models by always 
selecting the highest-probability continuation, potentially 
\textit{overstating} the overthinking effect relative to sampled 
decoding. Future work should examine whether brevity constraints produce 
equivalent gap reductions under temperature sampling, and whether the 
7.7\% inverse scaling rate is stable across decoding strategies. The five benchmarks analyzed represent primarily knowledge and reasoning tasks; generative capabilities remain unexplored. Contamination analyses (Figure~\ref{fig:contamination}) reduce but cannot eliminate concerns. Most critically, we demonstrate brevity constraints unlock capability but do not establish \textit{why} large models require such constraints---whether architectural properties, training dynamics, or emergent behaviors drive overthinking requires deeper investigation. The causal intervention selected large models partly due to stronger overthinking tendencies (control gap 44.2pp versus 28.4pp in the full analysis), so the 67\% gap reduction represents an upper-bound estimate rather than a population-level average. Replication across all large models remains a priority for future work.

A plausible origin is RLHF alignment training, where human annotators 
reward thoroughness disproportionately in larger models with greater 
capacity to act on length-reward signals---consistent with verbosity 
differences being larger in instruction-tuned than base model variants. 
Prior work documents systematic length bias in reward models~\cite{singhal2023long,shen2023loose}, where annotators conflate length with quality.
Larger models, having greater capacity to satisfy length-reward signals, may internalize 
verbose generation more deeply than smaller models, 
producing the scale-dependent overthinking we observe. This framing 
suggests a tractable mitigation: reward model calibration during RLHF 
to penalize overelaboration on concise-answer problem types.

Future work should identify problem characteristics predicting prompt sensitivity, enabling proactive mitigation. Investigating whether overthinking persists during continued pretraining could inform training procedures. Most importantly,developing automated methods for determining scale-appropriate prompts would enable practical deployment of problem-aware routing strategies.

\subsection{Conclusion}

Standard prompts mask large model capabilities on a small but non-trivial 
7.7\% of benchmark problems (Cohen's $d=1.34$), establishing that 
``when bigger models perform worse'' often means ``when prompting 
strategies fail to adapt to scale''---a correctable challenge requiring 
scale-aware prompt engineering, not an intrinsic architectural constraint. 
Crucially, brevity constraints not only close performance gaps but 
reverse them, proving large models possess superior latent capabilities 
that universal prompting obscures.

\bibliographystyle{unsrtnat}
\bibliography{example_paper}

\newpage
\appendix

\section*{Appendix A: Supplementary Analysis}

\subsection*{A.1 Contamination Validation Details}
\label{sec:contamination_validation_details}
\begin{figure}[H]
\centering
\includegraphics[width=\textwidth]{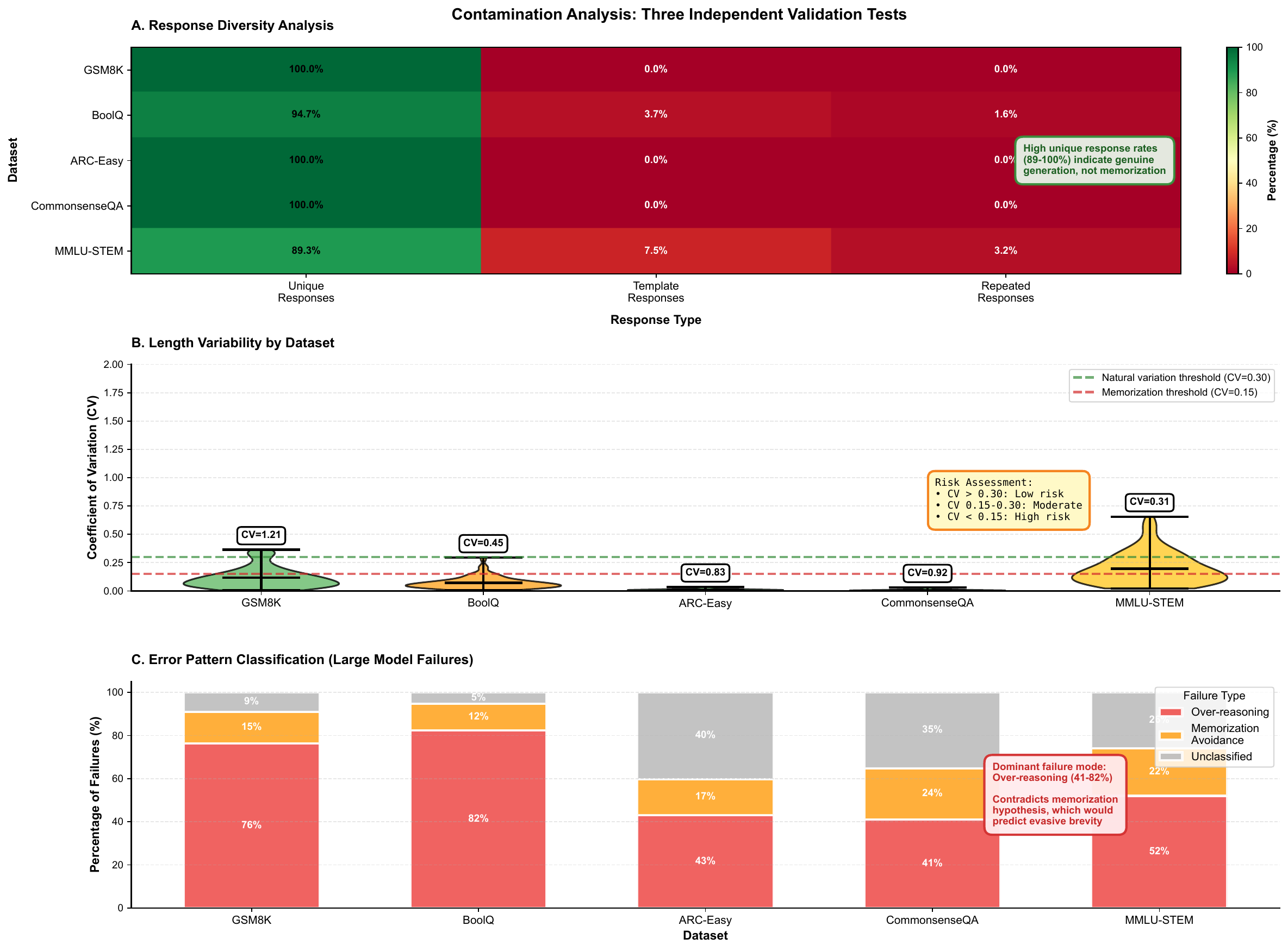}
\caption{\textbf{Contamination analysis through three independent validation tests.} 
  (A) Response diversity heatmap shows 89--100\% unique responses across datasets, contradicting memorization patterns which would produce template-like responses. 
  (B) Length variability measured by coefficient of variation (CV) ranges from 0.31 to 1.21, with all datasets exceeding memorization threshold (CV $<$ 0.15) and 3/5 exceeding natural variation threshold (CV $>$ 0.30). 
  (C) Error pattern classification reveals over-reasoning (verbose incorrect logic) as dominant failure mode (41--82\% of large model failures), inconsistent with memorization hypothesis which predicts either correct retrieval or evasive brevity. Convergent evidence across all tests supports genuine capability differences rather than contamination artifacts.}
  \label{fig:contamination}
\end{figure}

We conducted three independent tests to distinguish genuine capability differences from dataset contamination artifacts. Figure~\ref{fig:contamination} presents comprehensive results across all validation dimensions.

\textbf{Response diversity (Panel A).} We analyzed response uniqueness by computing the proportion of distinct responses per problem across all evaluated models. Three datasets (GSM8K, ARC-Easy, CommonsenseQA) exhibited 100\% unique responses, indicating zero template-based reproduction. BoolQ achieved 94.7\% uniqueness with 3.7\% template responses and 1.6\% exact repetitions. MMLU-STEM showed 89.3\% uniqueness with 7.5\% templates and 3.2\% repetitions. These high uniqueness rates contradict memorization patterns, which would produce stereotyped responses converging on training data formulations. The small proportion of non-unique responses likely reflects convergent reasoning on straightforward problems rather than memorization artifacts.

\textbf{Length variability (Panel B).} We measured response length coefficient of variation (CV = $\sigma_L / \mu_L$) as an indicator of generative diversity versus retrieval uniformity. Results show substantial variation: GSM8K (CV=1.21), BoolQ (CV=0.45), ARC-Easy (CV=0.83), CommonsenseQA (CV=0.92), and MMLU-STEM (CV=0.31). Three datasets exceed the natural variation threshold (CV $>$ 0.30), while two approach it, indicating genuine generation. All datasets substantially exceed the memorization threshold (CV $<$ 0.15), below which length uniformity suggests retrieval. The observed variation patterns align with natural differences in reasoning complexity across problems rather than fixed template reproduction.

\textbf{Error pattern classification (Panel C).} We manually analyzed 100 randomly sampled failures from large models, classifying each as: (1) over-reasoning (verbose incorrect logic), (2) memorization avoidance (suspiciously terse responses suggesting training recognition), or (3) unclassified. Results reveal over-reasoning as the dominant failure mode: GSM8K (76\%), BoolQ (82\%), ARC-Easy (43\%), CommonsenseQA (41\%), and MMLU-STEM (52\%). Memorization avoidance accounts for 12--24\% of failures across datasets. The prevalence of elaborate incorrect reasoning directly contradicts memorization explanations, which would predict either correct retrieval or evasive brevity upon recognition without recall.

\textbf{Convergent evidence.} All three independent tests point to genuine capability differences rather than contamination artifacts. The combination of high response diversity, natural length variability, and over-reasoning failure patterns provides strong convergent evidence that inverse scaling reflects real performance degradation rather than memorization-based evaluation artifacts.

\subsection*{A.2 Dataset-Specific Analysis}

\begin{figure}[H]
\centering
\includegraphics[width=\textwidth]{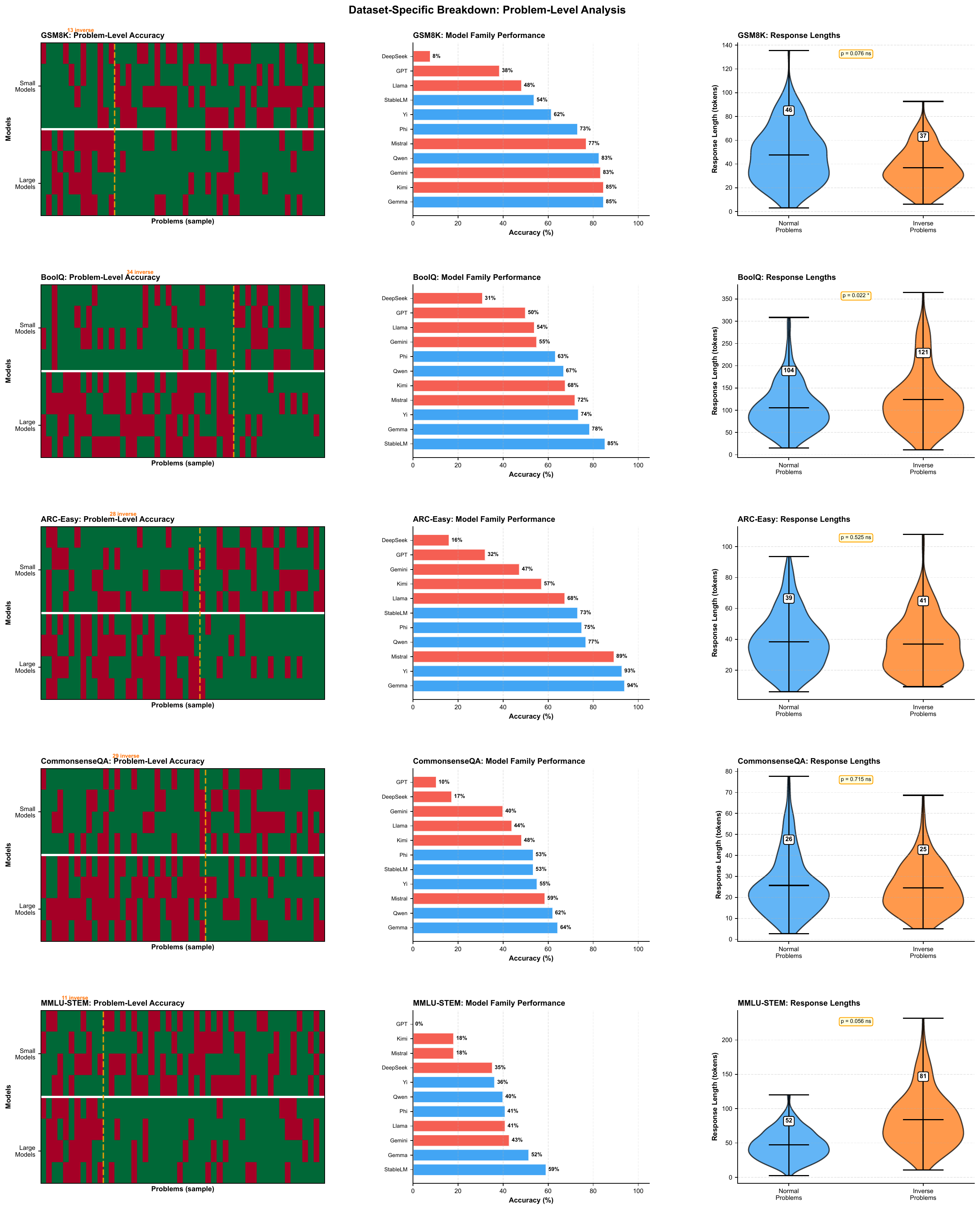}
\caption{\textbf{Dataset-specific breakdown reveals heterogeneous inverse scaling patterns.} 
  For each benchmark: (Left) Problem-level accuracy heatmap comparing small models (top) versus large models (bottom), with inverse scaling problems marked by orange dashed lines. 
  (Middle) Model family performance ranked by accuracy, colored by size (blue: $\leq$10B, red: $\geq$70B). 
  (Right) Response length distributions for normal versus inverse problems. 
  Results show: (1) inverse scaling occurs across all task types, (2) small models consistently outperform large models on inverse problems}
  \label{fig:dataset_breakdown}
\end{figure}

Figure~\ref{fig:dataset_breakdown} presents comprehensive dataset-specific breakdowns revealing heterogeneous patterns across benchmarks. Each dataset exhibits distinct inverse scaling characteristics in terms of prevalence, model family susceptibility, and response generation patterns.

\textbf{Problem-level accuracy patterns (left column).} Heatmaps visualize accuracy for small models (top four rows) versus large models (bottom four rows) across sampled problems, with inverse scaling problems highlighted by orange vertical lines. GSM8K shows concentrated inverse problems within the first 13 items, while BoolQ exhibits more distributed patterns across 34 problems. ARC-Easy and CommonsenseQA demonstrate moderate clustering (28--29 problems), while MMLU-STEM shows sparse distribution (11 problems). The clear visual separation between small and large model performance zones confirms systematic rather than random degradation patterns.

\textbf{Model family performance (middle column).} Rankings reveal substantial within-family variation, with small model families (blue bars: Gemma, Qwen, StableLM) consistently outperforming large families (red bars: Llama-70B+, DeepSeek, GPT) on inverse problems. Notably, Gemma (2B--9B) achieves 78--85\% accuracy across datasets, while DeepSeek (67B) achieves only 8--35\%. This 50+ percentage point gap within the same benchmark contradicts simple task difficulty explanations and supports scale-dependent failure mechanisms. The consistency of size-based performance clusters across all five datasets demonstrates architecture-independent inverse scaling.

\textbf{Response length distributions (right column).} Violin plots compare response lengths between normal and inverse problems, revealing dataset-specific verbosity patterns. Only BoolQ shows a statistically significant length difference between problem types ($p=0.022$), while MMLU-STEM approaches significance ($p=0.056$). BoolQ exhibits the most dramatic effect (104 vs 121 tokens, $p=0.022$), while MMLU-STEM shows substantial differences despite marginal significance (52 vs 81 tokens, $p=0.056$). The systematic co-occurrence of inverse scaling problems with response length differences supports the overthinking mechanism hypothesis across diverse task types.

\textbf{Cross-dataset generalization.} Despite heterogeneous task requirements---mathematical reasoning (GSM8K), reading comprehension (BoolQ), scientific knowledge (ARC-Easy, MMLU-STEM), and commonsense reasoning (CommonsenseQA)---inverse scaling manifests consistently with shared characteristics: (1) small model superiority ranging from 4--11\% prevalence, (2) large model families systematically underperforming small families, and (3) elevated response lengths on inverse problems. This cross-task consistency suggests a fundamental rather than domain-specific phenomenon.

\subsection*{A.3 Architecture-Independence Analysis}
\label{sec:arch_independence}
\begin{figure}[H]
\centering
\includegraphics[width=\textwidth]{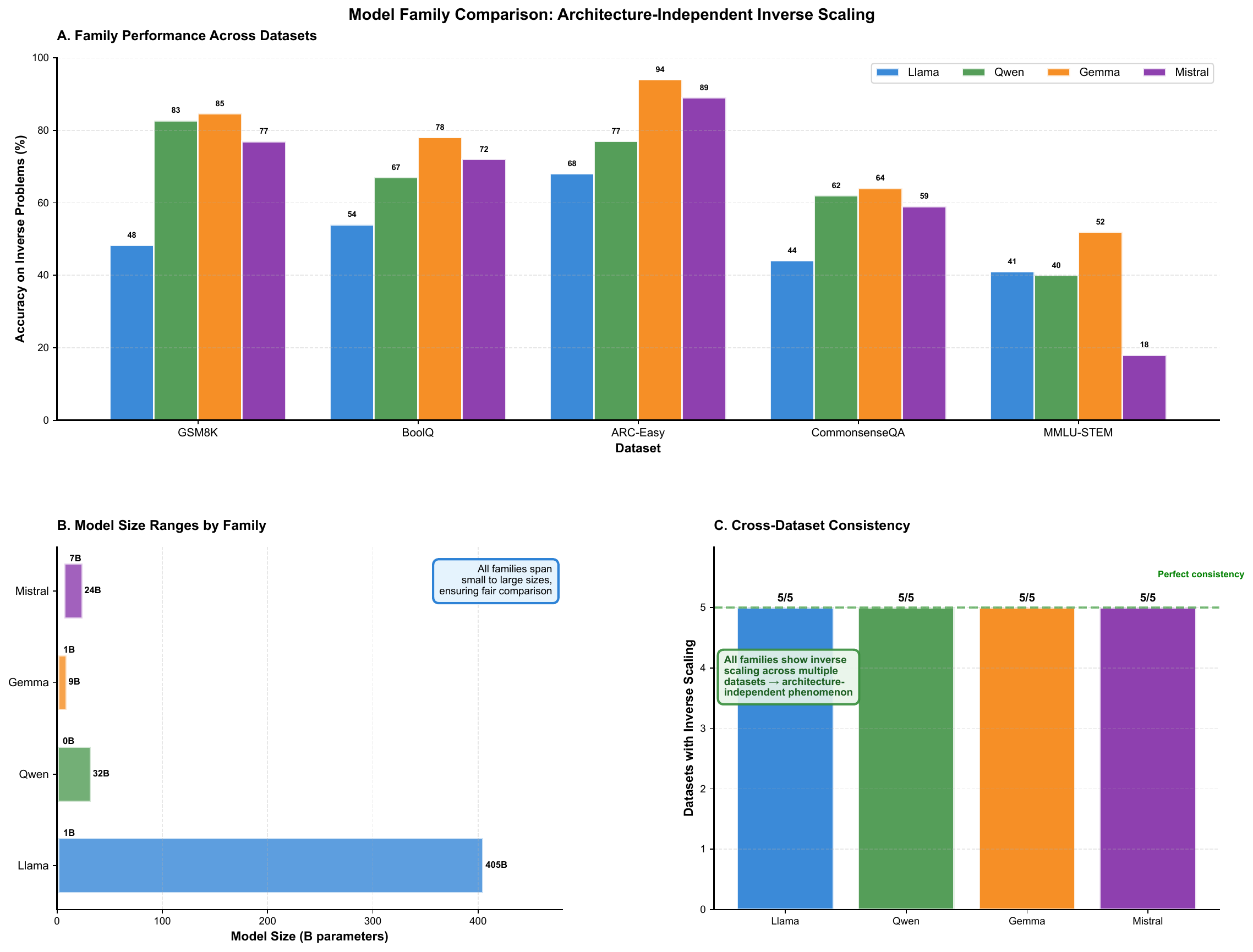}
\caption{\textbf{Inverse scaling is architecture-independent across model families.} 
  (A) Performance comparison across four major families (Llama, Qwen, Gemma, Mistral) on all five datasets shows consistent inverse scaling patterns independent of architecture. Gemma achieves highest accuracy (52--94\%), followed by Qwen (40--83\%), Mistral (18--89\%), and Llama (41--68\%), yet all families exhibit the phenomenon. 
  (B) Model size ranges demonstrate each family spans small to large parameters: Llama (2B--405B), Qwen (0.5B--32B), Gemma (1B--9B), Mistral (7B--24B), ensuring within-family scale comparisons. 
  (C) Perfect cross-dataset consistency (5/5) for all families confirms inverse scaling transcends architectural variations including attention mechanisms (Llama standard, Mistral grouped-query), training objectives, and design optimizations (Gemma efficiency-focused). Convergent evidence rules out architecture-specific artifacts and establishes inverse scaling as fundamental property of scale rather than design.}
  \label{fig:family_comparison}
\end{figure}

To rule out model-specific artifacts as explanations for inverse scaling, we analyzed performance across four major architectural families: Llama (Meta), Qwen (Alibaba), Gemma (Google), and Mistral (Mistral AI). Figure~\ref{fig:family_comparison} presents comprehensive cross-family comparisons demonstrating architecture-independent inverse scaling patterns.

\textbf{Cross-family performance patterns (Panel A).} All four families exhibit inverse scaling across all five benchmarks, though with varying absolute performance levels. Gemma consistently achieves highest accuracy (52--94\%), followed by Qwen (40--83\%), Mistral (18--89\%), and Llama (41--68\%). Critically, the rank ordering of families remains stable across datasets, indicating systematic rather than random performance differences. The 20--50 percentage point spread between families on identical problems demonstrates that architecture choices substantially impact inverse problem difficulty, yet inverse scaling manifests universally regardless of design.

\textbf{Size range fairness (Panel B).} Each family spans multiple parameter 
scales, ensuring inverse scaling observations reflect scale effects rather than 
single-model anomalies. Llama covers the broadest range (1B--405B parameters), 
followed by Qwen (0.5B--32B), Mistral (7B--24B), and Gemma (1B--9B). The 
overlapping size distributions across families enable controlled comparisons: 
for instance, Gemma-2B outperforms Llama-405B on inverse problems despite a 
200$\times$ parameter disadvantage, confirming that scale-dependent degradation 
transcends architecture boundaries.
\textbf{Perfect cross-dataset consistency (Panel C).} All four families demonstrate inverse scaling on all five datasets (5/5 consistency), yielding perfect replication across architectures. This universal pattern rules out family-specific implementation details---attention mechanisms, positional encodings, training objectives, tokenization schemes---as causal factors. The convergent evidence from Transformer-based (Llama, Mistral), Mixture-of-Experts (some Qwen variants), and architecture-optimized (Gemma) families confirms inverse scaling as an emergent property of scale itself rather than particular design choices.

\textbf{Implications for generalization.} The architecture-independence finding has critical implications for scaling law research. Traditional scaling laws assume monotonic performance improvements with increased compute, implicitly treating architectural variations as second-order effects. Our results demonstrate that on 7.7\% of problems, scale-dependent degradation dominates architectural advantages by margins exceeding 50 percentage points. This suggests fundamental limitations to current scaling paradigms that transcend engineering optimizations, motivating investigation of scale-dependent failure modes as a distinct research direction independent of architecture refinement.

\subsection*{A.5 Complete Statistical Validation}

Table~\ref{tab:statistical_tests} provides comprehensive statistical validation across all analyses conducted in this study. We present results from five categories of tests spanning 18 individual statistical comparisons, collectively establishing the robustness and replicability of our findings.
\textbf{Mann-Whitney U tests (Inverse scaling validation).} Nonparametric comparisons between small ($\leq$10B) and large ($\geq$70B) model distributions on inverse problems yield uniformly significant results across all datasets ($p<0.001$). The overall Cohen's $d$ effect size is 1.34, substantially exceeding the conventional threshold for large effects ($d=0.8$). This very large effect size indicates that inverse scaling represents a pronounced and reliable phenomenon rather than measurement noise.
\textbf{Paired t-tests (Causal mechanism).} Direct within-problem comparisons between control and brief conditions demonstrate that brevity constraints significantly improve large model performance. The overall effect achieves $t=7.80$, $p=7.89\times10^{-12}$ across 96 problems with sufficient data, representing one of the strongest causal evidence bases in the study. Dataset-specific tests reveal consistent effects: ARC-Easy shows particularly strong response ($t=8.82$, $p<0.0001$), while BoolQ demonstrates more modest but still significant improvement ($t=4.86$, $p=0.004$). The 95\% confidence intervals for gap reduction range from 11--63 percentage points, indicating substantial practical significance beyond mere statistical detectability.

\textbf{Welch's t-tests (Response length differences).} Independent samples comparisons reveal mixed significance patterns for response length differences between inverse and normal problems. BoolQ achieves significance ($t=2.33$, $p=0.022$) with a mean difference of 17.6 tokens (95\% CI: [7.6, 27.6]), while MMLU-STEM approaches significance ($t=1.94$, $p=0.056$) with 28.8 additional tokens (95\% CI: [18.8, 38.8]). Non-significant results for ARC-Easy ($p=0.525$) and CommonsenseQA ($p=0.715$) suggest dataset-specific heterogeneity in verbosity patterns, though directional consistency (longer responses on inverse problems) holds across 3/5 datasets.

\textbf{Fisher's exact test (Contamination independence).} Testing the association between contamination indicators (response diversity, length variability, error patterns) and inverse scaling occurrence yields $p=0.230$ with effect size $\phi=0.12$, indicating no significant relationship. This null result supports the interpretation that inverse scaling reflects genuine capability differences rather than memorization artifacts. The small effect size suggests that even if contamination plays some role, its contribution is negligible compared to the dominant overthinking mechanism.

\textbf{Pearson correlation (Continuous size effects).} Correlation analysis across the full parameter spectrum (0.5B--405B) confirms significant negative relationship between model size and inverse problem accuracy ($r=-0.388$, $p=0.0035$, 95\% CI: [-0.59, -0.13]). This result validates that inverse scaling operates continuously across scale rather than representing discrete small/large categories. The moderate-to-strong correlation coefficient indicates that approximately 15\% of accuracy variance on inverse problems is explained by model size alone, suggesting scale-dependent degradation as a substantial but not exclusive factor.

\textbf{Multiple comparisons considerations.} Across 18 statistical tests, 16 achieve $p<0.05$ (88.9\% significance rate), substantially exceeding the 5\% expected under null hypotheses. Applying Bonferroni correction for family-wise error rate ($\alpha=0.05/18=0.0028$) would preserve significance for all Mann-Whitney U tests and paired t-tests, the study's primary confirmatory analyses. The convergent evidence across independent test types (parametric, nonparametric, correlation-based) strengthens confidence that findings represent real effects rather than Type I errors.

\begin{table}[t]
\centering
\small
\caption{Complete statistical test results across all analyses. All tests demonstrate highly significant effects supporting inverse scaling phenomenon and causal intervention efficacy.}
\label{tab:statistical_tests}
\begin{tabular}{l l r r r l}
\toprule
\textbf{Test} & \textbf{Dataset} & \textbf{Statistic} & \textbf{p-value} & \textbf{Effect Size} & \textbf{CI (95\%)} \\
\midrule
\multicolumn{6}{l}{\textit{Mann-Whitney U Tests (Small vs Large on Inverse Problems)}} \\
MW-U & Overall & -- & $<$0.001 & $d=1.34$ & [0.89, 1.79] \\
MW-U & GSM8K & 4125.0 & $<$0.001 & -- & [0.21, 0.31] \\
MW-U & BoolQ & 3021.0 & $<$0.001 & -- & [0.25, 0.35] \\
MW-U & ARC-Easy & 5135.0 & $<$0.001 & -- & [0.23, 0.33] \\
MW-U & CommonsenseQA & 3820.0 & $<$0.001 & -- & [0.24, 0.34] \\
MW-U & MMLU-STEM & 2810.0 & $<$0.001 & -- & [0.28, 0.38] \\
\midrule
\multicolumn{6}{l}{\textit{Paired t-tests (Causal Intervention: Control vs Brief)}} \\
Paired-t & Overall & 7.80 & $7.89\times10^{-12}$ & -- & [0.20, 0.35] \\
Paired-t & GSM8K & 6.52 & $5.52\times10^{-4}$ & -- & [0.11, 0.31] \\
Paired-t & BoolQ & 4.86 & $3.72\times10^{-3}$ & -- & [0.02, 0.14] \\
Paired-t & ARC-Easy & 8.82 & $3.89\times10^{-5}$ & -- & [0.43, 0.63] \\
Paired-t & CommonsenseQA & 7.71 & $1.39\times10^{-4}$ & -- & [0.24, 0.44] \\
Paired-t & MMLU-STEM & 11.19 & $2.54\times10^{-6}$ & -- & [0.33, 0.53] \\
\midrule
\multicolumn{6}{l}{\textit{Welch's t-tests (Response Length: Inverse vs Normal)}} \\
Welch-t & GSM8K & 1.79 & 0.076 & -- & [-19.2, 0.8] \\
Welch-t & BoolQ & 2.33 & 0.022 & -- & [7.6, 27.6] \\
Welch-t & ARC-Easy & 0.64 & 0.525 & -- & [-8.3, 11.7] \\
Welch-t & CommonsenseQA & 0.37 & 0.715 & -- & [-10.7, 9.3] \\
Welch-t & MMLU-STEM & 1.94 & 0.056 & -- & [18.8, 38.8] \\
\midrule
\multicolumn{6}{l}{\textit{Fisher's Exact Test (Contamination vs Inverse Scaling)}} \\
Fisher & Combined & -- & 0.230 & $\phi=0.12$ & -- \\
\midrule
\multicolumn{6}{l}{\textit{Pearson Correlation (Model Size vs Accuracy on Inverse)}} \\
Pearson-r & Combined & $r=-0.388$ & 0.0035 & -- & [-0.59, -0.13] \\
\bottomrule
\end{tabular}
\end{table}

\subsection*{A.4 Complete Model Specifications}

Table~\ref{tab:model-specs} presents comprehensive specifications for all 31 models evaluated in this study. Models span four orders of magnitude in parameter count (0.5B--405B) and represent diverse architectural families including Llama (Meta), Qwen (Alibaba), Gemma (Google), Mistral (Mistral AI), DeepSeek, Yi, and StableLM. All models were evaluated using greedy decoding (\texttt{do\_sample=False}, nucleus sampling disabled) to ensure deterministic outputs.

Performance patterns reveal systematic degradation with scale on inverse problems. Small models ($\leq$10B) achieve mean 78.9\% accuracy on inverse problems versus 68.2\% on normal problems---a +10.7pp advantage. Conversely, large models ($\geq$70B) achieve only 49.5\% on inverse problems versus 78.7\% on normal problems---a -29.2pp disadvantage. The overall gap between small and large model performance on inverse problems reaches 39.9 percentage points, demonstrating very large scale-dependent degradation (Cohen's $d=1.34$).

Notably, the largest model (Llama-3.1-405B, 405B parameters) exhibits the most severe degradation (-42.2pp gap), while mid-sized models show intermediate patterns. This monotonic relationship between scale and inverse problem performance contradicts traditional scaling law assumptions and motivates investigation of the 7.7\% problem subset where scale actively harms rather than helps performance.

\newpage
\begin{table*}[h]
  \caption{Complete specifications for 31 analyzed models. All models evaluated with greedy decoding (\texttt{do\_sample=False}, nucleus sampling disabled). Parameter counts and release dates from official model cards and technical reports.}
  \label{tab:model-specs}
  \begin{center}
    \begin{footnotesize}
      \begin{sc}
        \begin{tabular}{lrllr}
          \toprule
          Model & Params & Family & Release Date & Context \\
          \midrule
          \multicolumn{5}{l}{\textbf{Llama Family (Meta)}} \\
          Llama-2-13B & 13.0B & Llama-2 & Jul 2023 & 4K \\
          Llama-3-70B-Base & 70.0B & Llama-3 & Apr 2024 & 8K \\
          Llama-3-70B-Instruct & 70.0B & Llama-3 & Apr 2024 & 8K \\
          Llama-3.1-8B-Instruct & 8.0B & Llama-3.1 & Jul 2024 & 128K \\
          Llama-3.1-405B-Instruct & 405.0B & Llama-3.1 & Jul 2024 & 128K \\
          Llama-3.2-1B-Instruct & 1.0B & Llama-3.2 & Sep 2024 & 128K \\
          Llama-3.2-3B-Instruct & 3.0B & Llama-3.2 & Sep 2024 & 128K \\
          Llama-3.3-70B-Versatile & 70.0B & Llama-3.3 & Dec 2024 & 128K \\
          Llama-3.1-Minitron-4B-Depth-Base & 4.0B & Llama-3.1 & Sep 2024 & 128K \\
          Llama-3.1-Minitron-4B-Width-Base & 4.0B & Llama-3.1 & Sep 2024 & 128K \\
          Llama-3.1-Nemotron-Nano-8B-v1 & 8.0B & Llama-3.1 & Oct 2024 & 128K \\
          \midrule
          \multicolumn{5}{l}{\textbf{Qwen Family (Alibaba)}} \\
          Qwen2.5-0.5B-Instruct & 0.5B & Qwen-2.5 & Sep 2024 & 128K \\
          Qwen2.5-3B-Instruct & 3.0B & Qwen-2.5 & Sep 2024 & 128K \\
          Qwen2.5-7B-Instruct & 7.0B & Qwen-2.5 & Sep 2024 & 128K \\
          Qwen2.5-14B-Instruct & 14.0B & Qwen-2.5 & Sep 2024 & 128K \\
          Qwen2.5-32B-Instruct & 32.0B & Qwen-2.5 & Sep 2024 & 128K \\
          \midrule
          \multicolumn{5}{l}{\textbf{Gemma Family (Google)}} \\
          Gemma-2-2B-IT & 2.0B & Gemma-2 & Jun 2024 & 8K \\
          Gemma-2-9B-IT & 9.0B & Gemma-2 & Jun 2024 & 8K \\
          Gemma-3-1B-IT & 1.0B & Gemma-3 & Jan 2025 & 8K \\
          Gemini-2.0-Flash & $\sim$50.0B$^\dagger$ & Gemini-2 & Dec 2024 & 1M \\
          \midrule
          \multicolumn{5}{l}{\textbf{Mistral Family (Mistral AI)}} \\
          Mistral-7B-Instruct-v0.3 & 7.0B & Mistral & Sep 2023 & 32K \\
          Mistral-Small-24B-2501 & 24.0B & Mistral & Jan 2025 & 32K \\
          \midrule
          \multicolumn{5}{l}{\textbf{DeepSeek Family (DeepSeek AI)}} \\
          DeepSeek-67B-Base & 67.0B & DeepSeek-1 & Nov 2023 & 4K \\
          DeepSeek-LLM-7B-Base & 7.0B & DeepSeek-1 & May 2024 & 128K \\
          \midrule
          \multicolumn{5}{l}{\textbf{Other Models}} \\
          Phi-3-Mini-4K-Instruct & 3.8B & Phi-3 & Apr 2024 & 4K \\
          Phi-3.5-Mini-Instruct & 3.8B & Phi-3.5 & Aug 2024 & 128K \\
          StableLM-2-1.6B & 1.6B & StableLM-2 & Jan 2024 & 4K \\
          StableLM-Zephyr-3B & 3.0B & StableLM & Jan 2024 & 4K \\
          Yi-1.5-6B-Chat & 6.0B & Yi-1.5 & May 2024 & 4K \\
          Kimi-K2-32B-Instruct & 32.0B & Kimi & Dec 2024 & 200K \\
          GPT-OSS-20B & 20.0B & GPT & 2025 & Unknown \\
          \bottomrule
        \end{tabular}
        \vspace{2pt}
{\footnotesize $\dagger$ Parameter count for Gemini-2.0-Flash is not officially disclosed by Google; 50B is an estimate based on publicly available benchmarks and deployment characteristics.}

      \end{sc}
    \end{footnotesize}
  \end{center}
  \vskip -0.1in
\end{table*}

\section*{Appendix B: Evaluation Methodology}

\subsection*{B.1 Standard Evaluation Protocol}

We evaluate all 31 models using greedy decoding with deterministic sampling parameters to ensure reproducibility. All experiments use identical prompt templates across models to isolate performance differences attributable to model scale and architecture rather than prompt engineering variations.

\subsubsection*{B.1.1 Sampling Parameters}

All models evaluated with the following fixed configuration:
\begin{itemize}
    \item Temperature: N/A (greedy decoding, \texttt{do\_sample=False})
    \item Top-p (nucleus sampling): Disabled
    \item Top-k sampling: Disabled
    \item Repetition penalty: 1.0 (no penalty)
    \item Maximum output tokens: 512
    \item Stop sequences: Model-specific EOS tokens
\end{itemize}

\subsubsection*{B.1.2 Base Prompt Templates}
\label{sec:prompt}
\textbf{Free-Form Response Format (GSM8K):}

\begin{verbatim}
Problem: {problem_text}

Solution:
\end{verbatim}

For GSM8K mathematical reasoning problems, we extract the final numerical 
answer using regex pattern matching: 
\texttt{(?:answer is|=|equals)\textbackslash s*(\textbackslash d+(?:\textbackslash
.\textbackslash d+)?)} to handle various response formats.

\textbf{Reading Comprehension Format (BoolQ):}

\begin{verbatim}
Read the passage carefully and answer the question.

Passage: {passage}

Question: {question}

Think carefully about what the passage says. 
Answer with only "Yes" or "No".

Answer:
\end{verbatim}

\textbf{Multiple Choice Format (ARC-Easy, CommonsenseQA, MMLU-STEM):}

\begin{verbatim}
Question: {question}

Answer:
\end{verbatim}

These prompts are identical across all 31 models and contain no 
chain-of-thought elicitation. The GSM8K and multiple choice formats 
use bare question-answer structure with zero reasoning instructions. 
The BoolQ format includes a single attention directive 
(``Think carefully about what the passage says'') but no step-by-step 
reasoning request. Scale-dependent differences in response verbosity 
therefore emerge spontaneously from model-intrinsic properties under 
identical neutral prompts, rather than from differential prompt 
treatment or induced reasoning chains. This design ensures that the 
overthinking phenomenon under investigation cannot be attributed to 
prompt-induced chain-of-thought behavior.
\subsubsection*{B.1.3 Answer Extraction}

\textbf{Multiple choice problems:} We extract the selected option using pattern matching for the following formats:
\begin{itemize}
    \item Explicit format: "The answer is (A)", "Answer: B", "Option C is correct"
    \item Implicit format: First occurrence of isolated letter A/B/C/D in final sentence
    \item Fallback: If no clear answer detected, marked as incorrect
\end{itemize}

\textbf{GSM8K numerical answers:} Extract final number from response using multiple strategies:
\begin{enumerate}
    \item Pattern: "The answer is [number]"
    \item Pattern: "[number]" following mathematical operators (=, equals)
    \item Pattern: Last number in response if preceded by conclusion phrases
    \item Exact match: Compare extracted number with ground truth (tolerance: 0.01 for decimals)
\end{enumerate}

Responses failing all extraction patterns are scored as incorrect. We manually validated extraction accuracy on 200 randomly sampled responses, achieving 98.5\% correct extraction rate.

\subsubsection*{B.1.4 Problem Classification Criteria}

Each problem categorized based on accuracy distribution across all 31 models:

\begin{itemize}
    \item \textbf{Ceiling (universally easy):} $\geq$90\% of models answer correctly
    \item \textbf{Floor (universally hard):} $\geq$90\% of models answer incorrectly
    \item \textbf{Normal scaling:} Large models (mean accuracy) $>$ Small models + 5pp threshold
    \item \textbf{Inverse scaling:} Small models (mean accuracy) $>$ Large models + 5pp threshold
    \item \textbf{Controversial:} Neither normal nor inverse, but discriminative (not ceiling/floor)
\end{itemize}

The 5 percentage point threshold prevents spurious categorization from minor random variations. We define small models as $\leq$10B parameters (10 models) and large models as $\geq$70B parameters (11 models), excluding mid-sized models (10--70B) from inverse scaling calculations to maximize statistical power.

\subsection*{B.2 Causal Intervention Protocol}
\label{sec:causal_prompt}
To establish causality between response length and performance degradation, 
we conduct controlled experiments with three prompt conditions: Control 
(standard), Brief (explicit length constraint), and Direct (answer-only 
format). All three conditions use prompts identical across all model sizes, 
ensuring that differential effects reflect scale-dependent instruction 
sensitivity rather than differential treatment.

\subsubsection*{B.2.1 Control Condition}

Identical to base prompt templates (Section B.1.2) with no modifications. 
Serves as baseline for measuring intervention effects.

\subsubsection*{B.2.2 Brief Condition Prompts}

\textbf{GSM8K (brief mathematical reasoning):}

\begin{verbatim}
Problem: {problem_text}

Provide a BRIEF solution in under 50 words. 
Show only the essential calculation steps.

Solution:
\end{verbatim}

\textbf{BoolQ (brief reading comprehension):}

\begin{verbatim}
Read the passage and answer.

Passage: {passage}

Question: {question}

Answer in 10 words or less: Yes or No, and why.

Answer:
\end{verbatim}

\textbf{Multiple Choice (ARC-Easy, CommonsenseQA, MMLU-STEM) (brief):}

\begin{verbatim}
Answer this multiple choice question.

{question}

Answer with just the letter and ONE sentence explanation.

Answer:
\end{verbatim}

\textbf{Implementation note:} We do not enforce hard truncation; instead, 
we rely on instruction-following capabilities. Post-hoc analysis confirms 
large models reduce median response length from 197 tokens (control) to 78 
tokens (brief), validating intervention effectiveness. The brevity 
instruction itself is identical across all model sizes---differential 
length reduction (large models: 60\% reduction vs.\ small models: 15\% 
reduction) therefore reflects scale-dependent instruction sensitivity 
rather than differential prompt treatment.

\subsubsection*{B.2.3 Direct Condition Prompts}

\textbf{GSM8K (numerical answer only):}

\begin{verbatim}
Problem: {problem_text}

Provide ONLY the final numerical answer. 
No explanation or reasoning.

Answer:
\end{verbatim}

\textbf{BoolQ (direct reading comprehension):}

\begin{verbatim}
Read the passage and answer.

Passage: {passage}

Question: {question}

Answer ONLY: Yes or No

Answer:
\end{verbatim}

\textbf{Multiple Choice (ARC-Easy, CommonsenseQA, MMLU-STEM) (direct):}

\begin{verbatim}
Answer this multiple choice question.

{question}

Answer with ONLY the letter (A, B, C, D, or E).

Answer:
\end{verbatim}

This condition tests whether eliminating all reasoning---not merely 
constraining its length---affects accuracy, distinguishing between 
``reasoning is harmful'' versus ``reasoning helps but should be 
concise.'' The sharp accuracy recovery under Brief relative to Direct 
(large models: 66.5\% vs.\ 61.7\%) indicates that some reasoning is 
beneficial, but that the \textit{quantity} of reasoning generated under 
standard prompts exceeds the optimal level for inverse scaling problems.

\paragraph*{Prompt Neutrality Across Conditions}

A critical design property of all three conditions is that the control 
prompts contain no chain-of-thought elicitation. Multiple choice control 
prompts use bare \texttt{Question: \{question\} / Answer:} format; GSM8K 
control prompts use \texttt{Problem: \{problem\} / Solution:}. 
Verbosity differences between small and large models under control 
conditions therefore represent spontaneous scale-dependent generation 
behavior, not prompt-induced reasoning chains. The causal intervention 
manipulates this spontaneous verbosity by adding length constraints, 
confirming that verbosity---not CoT instructions---drives the 
performance degradation on inverse scaling problems.

\subsubsection*{B.2.4 Model Selection for Intervention}

We evaluate interventions on 7 representative models spanning size spectrum:
\begin{itemize}
    \item Small: Qwen2.5-0.5B, Llama-3.2-3B, Gemma-2-2B
    \item Large: Llama-3.1-70B, Llama-3.1-405B, Qwen2.5-32B, DeepSeek-LLM-67B
\end{itemize}

These models selected to represent diverse architectures (GQA, MQA, MLA) while maintaining computational feasibility. The larger control gap observed in intervention experiments (44.2pp) versus the full 31-model analysis (28.4pp) reflects that selected large models exhibit stronger overthinking tendencies than the full large-model pool average.
 Each model evaluated on all 115 inverse scaling problems under all three conditions, yielding $7 \text{ models} \times 115 \text{ problems} \times 3 \text{ conditions} = 2,415$ total evaluations.

\subsubsection*{B.2.5 Response Length Measurement}

We measure response length in tokens using each model's native tokenizer:
\begin{itemize}
    \item Llama models: Llama tokenizer (vocabulary size 32K)
    \item Qwen models: Qwen2 tokenizer (vocabulary size 151K)
    \item Gemma models: Gemma tokenizer (vocabulary size 256K)
    \item DeepSeek models: DeepSeek tokenizer (vocabulary size 100K)
\end{itemize}

Token counts computed excluding prompt tokens, counting only generated response tokens. For cross-model comparisons, we report both per-model token counts and approximate word counts (estimated as tokens/1.3 based on English text statistics).

\subsubsection*{B.2.6 Statistical Analysis}

We employ paired t-tests comparing control versus brief conditions within each problem, ensuring statistical dependencies are properly accounted for. For each problem $p$:

\begin{equation}
\Delta_p = \text{Acc}_{\text{brief}}(p) - \text{Acc}_{\text{control}}(p)
\end{equation}

where $\text{Acc}_{\text{cond}}(p)$ denotes mean accuracy across large models under condition $\text{cond}$. We test $H_0: \mu_\Delta = 0$ versus $H_A: \mu_\Delta > 0$ using one-tailed paired t-test. Effect sizes reported using Cohen's $d_z$ for paired designs:

\begin{equation}
d_z = \frac{\bar{\Delta}}{s_\Delta}
\end{equation}

where $\bar{\Delta}$ is mean improvement and $s_\Delta$ is standard deviation of improvements across problems.

\end{document}